# Teach and Repeat Navigation: A Robust Control Approach

Payam Nourizadeh, Michael Milford and Tobias Fischer

*Abstract*—Robot navigation requires an autonomy pipeline that is robust to environmental changes and effective in varying conditions. Teach and Repeat (T&R) navigation has shown high performance in autonomous repeated tasks under challenging circumstances, but research within T&R has predominantly focused on *motion planning* as opposed to *motion control*. In this paper, we propose a novel T&R system based on a robust motion control technique for a skid-steering mobile robot using sliding-mode control that effectively handles uncertainties that are particularly pronounced in the T&R task, where sensor noises, parametric uncertainties, and wheel-terrain interaction are common challenges. We first theoretically demonstrate that the proposed T&R system is globally stable and robust while considering the uncertainties of the closed-loop system. When deployed on a Clearpath Jackal robot, we then show the global stability of the proposed system in both indoor and outdoor environments covering different terrains, outperforming previous state-of-the-art methods in terms of mean average trajectory error and stability in these challenging environments. This paper makes an important step towards long-term autonomous T&R navigation with ensured safety guarantees.

## I. INTRODUCTION

Mobile robots necessitate a comprehensive autonomy pipeline, encompassing perception, localization, path planning and motion control, to autonomously navigate different environments [1], [2]. However, assembling all the necessary components for autonomous navigation in diverse conditions poses challenges, including issues related to sensor integration, environmental variability, and hardware constraints. To address these challenges, Teach and Repeat (T&R), comprising two phases, teach and repeat, is a popular choice that has seen widespread adaptation [3]–[11].

During the teaching phase, the robot is manually or autonomously driven along a desired path while capturing the sensory information, which provides the path-planning and localization parts for the automation pipeline. In the repeat phase, the motion planning (perception) and motion control systems are used to follow the desired path from the teach phase under different environmental conditions. So far, the majority of research in the T&R domain has focused on the performance and robustness of the motion planning system [3], [7], [8], [11].

In this paper, we focus on the motion control system that needs to consider 1) possible changes in the dynamics of the robot due to the wheel-terrain interaction, the defined manoeuvre, and payload, 2) measurement and correction noises from both on-board sensors and the perception system, and 3) the uncertainties of the working environment,

The authors are with the QUT Centre for Robotics, Queensland University of Technology, Brisbane, QLD 4000, Australia (e-mail: payam.norizadeh@qut.edu.au). This work received funding from the Australian Government, via grant AUSMURIB000001 associated with ONR MURI grant N00014-19-1-2571 and an ARC Laureate Fellowship FL210100156 to MM. The authors acknowledge continued support from the Queensland University of Technology through the Centre for Robotics.

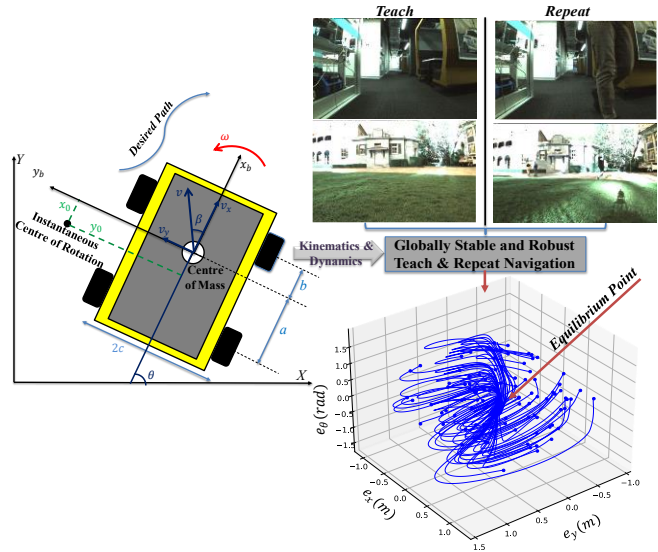

Figure 1. Our proposed approach for teach-and-repeat navigation relies on sliding-mode control techniques using wheel odometry and a low-resolution monocular camera. This combination is designed to deliver globally stable and robust navigation starting from any arbitrary initial tracking errors, even in challenging environments where terrain conditions and uncertainties can impact the system's dynamics.

especially in outdoor and off-road environments. Furthermore, since the kinematics equations of most mobile robots are nonlinear and subject to nonholonomic constraints, the trajectory tracking controller must employ nonlinear control techniques. Previous research in this field either designed linear controllers [11] or did not consider the robot's equations of motion and kinematics constraints [7], [8], hindering accurate and robust path following.

In this paper, as illustrated in Figure 1, we propose a novel, robust T&R system that is reliable and repeatable. Sliding mode control is used to design the motion controller based on the kinematics and dynamics equations of a skid-steering mobile robot. The proposed controller considers the parameter uncertainties of the system and guarantees the global stability of the closed-loop system. The inputs to our system are the wheel odometry and path corrections based on a motion planner operation on low-resolution monocular imagery [8]. Employing Sliding Mode Control (SMC) for visual T&R navigation is advantageous as it can effectively address the long-term autonomy challenge, where vision sensors are subject to disturbances from robot vibrations, changing lighting conditions, motion blur, and the presence of moving objects or humans in cluttered environments. The main contributions of this paper can be summarized as follows:

1. This paper proposes a robust T&R framework using sliding mode control that can operate under uncertainties, measurement noises, changes in the instantaneous centre of rotation location, and the wheel-terrain interaction in off-road terrains. The framework's robustness enhances the reliability and adaptability of the closed-loop system, especially in real-world environments.

2. The proposed framework guarantees the global stability of the closed-loop system, providing a sound theoretical foundation for the long-term autonomy challenge in T&R. As a result, as opposed to prior work [8], this controller does not require manual tuning for different types of manoeuvres and environments.

3. This paper evaluates the performance of the proposed framework in indoor and outdoor terrains on the Clearpath Jackal robot. These evaluations demonstrate the framework's generality and robustness in handling environmental uncertainties, showcasing its practical applicability in comparison with the state-of-the-art.

To support future research endeavours, we have made our code accessible for research purposes: https://github.com/QVPR/teach-repeat.

II. RELATED WORKS

T&R is one of the basic navigation techniques for mobile robots where the localization systems are not available either due to cost limitations or their operating environment, such as GPS-denied environments, underwater or in densely built-up cities [3], [9], [12]. T&R systems require both proprioceptive and exteroceptive sensors to measure the internal parameters of the robot and capture the information from the working environment during both the teach and repeat phases. Common proprioceptive sensors for T&R navigation are the wheel encoder and IMU to measure/estimate the linear and angular velocities and the position and orientation of the robot. Cameras are common exteroceptive sensors due to their low cost, availability, and capability to operate indoors and outdoors [3], [10]. These sensors not only provide visual information for the path-planner algorithm but also correct wheel odometry drifts and cumulative errors of the proprioceptive sensors [4].

*A. Motion-Planning*

Visual T&R navigation can be categorized into position-based or appearance-based approaches based on their mapping techniques [3], [8], [11]. Position-based approaches attempt to build a geometric map of the environment during the teaching phase [13], [14]. On the other hand, appearance-based approaches capture the visual information without building a map during the teaching phase [4], [8], [15], [16].

Research in the T&R domain primarily focused on motion planning, as opposed to motion control, and did not consider the nonlinear kinematics equations of wheeled mobile robots and their nonholonomic constraints [7], [8], [11], [13], [17], [18]. Therefore, without having a sufficiently accurate model of the robot, the research and analysis on the robustness and the stability of those T&R navigation systems can only be interpreted for the motion planning part rather than the entire closed-loop navigation system [1].

Dall'Osto et al. [8] proposed a low-cost bio-inspired T&R technique using wheel odometry and a low-resolution monocular camera. They used image processing techniques to rectify the drift of the wheel odometry by finding the similarity between the captured images during the teach and repeat phases. They investigated the robustness of their image processing technique to the different levels of wheel odometry corruptions. However, their controller [2] did not consider the exact kinematics equations of the robot and was only locally stable and highly reliant on the choice of hyperparameters and linear and angular velocity adjustments for each manoeuvre. In this paper, we use their low-resolution and fast vision technique as the motion planner.

Krajnik et al. [11] proposed a monocular T&R navigation system that uses visual information to correct the heading of the robot. They demonstrated the stability of their proposed navigation system based on a linear representation of the kinematics of Skid-Steering Mobile Robots (SSMR). However, the motion equations of SSMR are nonlinear and it is not a viable assumption to linearize their kinematics model as there is no fixed working point for these mobile systems. Rozsypálek et. al [4] designed a multidimensional particle filter to estimate the robot state in T&R navigation. Their technique estimates the longitudinal, lateral, and heading deviation of the robot in the repeat phase using visual information.

Furgale and Barfoot [7] proposed a visual T&R technique for mobile robots' navigation without requiring a localization system in off-road terrains. Clement et al. [6] introduced a robust T&R navigation method relying only on a monocular camera and conducted large-scale testing in off-road terrains. Based on a system control point of view, their technique addresses the path-planner section for the robot's automation pipeline and provides the desired values for the trajectory tracking system. However, a robust trajectory tracking system is required to steer the robot toward the desired trajectory in challenging conditions, especially in off-road terrains where wheel-terrain interaction could cause slippage and undesired skidding [19], [20].

*B. Motion-Control*

The design of motion controllers for wheeled mobile robots relies on the drive mechanism of the robot. In this paper, we use SSMRs as they are lightweight and suitable for both indoor and outdoor environments and focus on relevant trajectory tracking techniques [21]–[23]. Since skidding is the steering mechanism for these robots, tracking curvilinear trajectories for these robots is a challenging control task. Moreover, the location of the instantaneous centre of rotation (see Figure 1) is a parametric uncertainty for these robots [21], which makes the design of the controller complicated.

Due to the nonlinearity of the SSMR's kinematics equations, prior works mainly used nonlinear control techniques [24]–[26] to stabilize and regulate the tracking errors for this type of robot, including Lyapunov-based controllers [24], nonlinear model predictive control [27]–[29], backstepping technique [30], and robust control techniques [25], [31]. The main benefit of using nonlinear controllers for SSMR is demonstrating the stability of the closed-loop system and ensuring controllable behaviour for defined manoeuvres. However, only robust control techniques can consider unmodeled parameters and uncertainties, ensuring stability and long-term autonomy.

SMC is a well-established robust controller [32] and has shown reliable performance in solving different control problems for both holonomic [33]–[35] and nonholonomic [25], [31], [36] robots. SMC can stabilize the closed-loop system under disturbances and parametric uncertainties [37], which makes it a suitable candidate for SSMRs in both indoor and outdoor terrains, and T&R navigation systems.

Robust control techniques such as SMC can address the long-term autonomy challenge in T&R navigation. As discussed in Section II-A, vision-based T&R navigation relies on the vision sensors and these sensors are subject to unmodelled noises and disturbances due to the vibration of the robot, changes in lighting conditions, and motion blur (especially in low-light conditions). Additionally, the presence of moving objects or humans in cluttered environments can introduce uncertainty in image processing. The motivation for employing SMC lies in its ability to effectively mitigate these challenges, enhancing the robustness and reliability of T&R systems in real-world conditions. However, SMC suffers from the impractical chattering problem in actuators [25], [31], [38] and might be subject to singularity [31], [39], which needs to be addressed in the controller design procedure.

In this paper, we propose a robust T&R system using a combination of SMC and the appearance-based path planner developed by Dall'Osto et al. [8]. We design a globally stable and robust T&R navigation system considering the uncertainties of the closed-loop system and evaluate its performance in indoor and off-road terrains.

III. PROPOSED APPROACH

A. Image processing

We briefly review the image processing technique proposed by Dall'Osto et al. [8] as we used it as the motion planning component within our T&R navigation system. However, we note that our proposed motion controller is not tied to this particular motion planner.

The robot is manually driven on a desired path during the teaching phase. The odometry information and images are stored to generate a topometric map whenever the robot's displacement exceeds a certain threshold from the previous pose until reaches the final point. All images are converted to greyscale and patch normalization is applied [40].

In the repeat phase, the robot attempts to follow the desired trajectory relying on the same odometry readings and capturing images whilst traversing. The image processing technique compares query (repeat phase) with reference (teach phase) images and generates two correction offsets, i.e., in orientation and along-path displacement, to update the desired path in real-time. These offsets are computed based on the normalized cross-correlation, and we refer the interested reader to [8] for details.

B. Motion Control

This section describes the design of the motion controller which guides the robot along the desired path generated by the path planning system. Figure 1 shows an SSMR in a 2-D plane $(X, Y)$ with a local coordinate system $(x_b, y_b)$. The position and orientation of the robot's centre of mass in the global coordinate system is represented by $q = [x, y, \theta]^T$. $v$ and $\omega$ are linear and angular velocities of the robot (with respect to the centre of mass), and $v_x$ and $v_y$ are the longitudinal and lateral components of the robot's linear velocity in the local frame. $x_0$ and $y_0$ are longitudinal and lateral components of the robot's instantaneous centre of rotation with respect to the robot's local frame. The kinematics equations are as follows [22]:

$$\dot{q} = \begin{bmatrix} \dot{x} \\ \dot{y} \\ \dot{\theta} \end{bmatrix} = \begin{bmatrix} \cos\theta & x_0 \sin\theta \\ \sin\theta & -x_0 \cos\theta \\ 0 & 1 \end{bmatrix} \begin{bmatrix} v_x \\ \omega \end{bmatrix}, \quad (1)$$

where $v_x = \frac{r}{2}(\omega_R + \omega_L)$ and $\omega = \frac{r}{2c}(\omega_R - \omega_L)$, $\omega_R$ and $\omega_L$ are the angular velocities of the right and left wheels neglecting slipping and undesired skidding of the robot (more on the robot's slipping and undesired skidding in [31]). The dynamic model of the robot can be represented as follows [22]:

$$\begin{bmatrix} \dot{v}_x \\ \dot{\omega} \end{bmatrix} = \begin{bmatrix} \frac{c_3}{c_1}\omega^2 - \frac{c_4}{c_1}v_x \\ -\frac{c_5}{c_2}v_x\omega - \frac{c_6}{c_2}\omega \end{bmatrix} + \begin{bmatrix} \frac{1}{c_1}v_x^r \\ \frac{1}{c_2}\omega^r \end{bmatrix}, \quad (2)$$

where $c_1 : c_6$ are functions of some physical parameters of the robot, such as the mass, moment inertia, motor parameters, etc. [25], [31], and $v_x^r$ and $\omega^r$ are the reference control inputs that will be the output of the SMC.

To design the motion controller, first, we need to define the tracking errors dynamic. The tracking errors $q^e$ are defined based on the difference between the desired trajectory $q^d$ and $q$ as follows:

$$q^e = \begin{bmatrix} e_x \\ e_y \\ e_\theta \end{bmatrix} = \begin{bmatrix} x^d - x \\ y^d - y \\ \theta^d - \theta \end{bmatrix}. \quad (3)$$

The above error equation in the robot's local frame $q^\varepsilon$ and its dynamic $\dot{q}^\varepsilon$ are as follows:

$$q^\varepsilon = \begin{bmatrix} \varepsilon_1 \\ \varepsilon_2 \\ \varepsilon_3 \end{bmatrix} = \begin{bmatrix} \cos\theta & \sin\theta & 0 \\ -\sin\theta & \cos\theta & 0 \\ 0 & 0 & 1 \end{bmatrix} q^e \quad (4)$$

$$\dot{q}^\varepsilon = \begin{bmatrix} \omega\varepsilon_2 + v_x^d \cos\varepsilon_3 + \omega^d x_0 \sin\varepsilon_3 - v_x \\ (x_0 - \varepsilon_1)\omega + v_x^d \sin\varepsilon_3 - \omega^d x_0 \cos\varepsilon_3 \\ \omega^d - \omega \end{bmatrix}. \quad (5)$$

Now, we need to define the sliding surfaces for SMC [1], [25], [31], [38]. The objective of the trajectory tracking controller is chosen to regulate the position tracking errors ($\varepsilon_1$ and $\varepsilon_2$). Therefore, the sliding surfaces are defined as follows:

$$s_i = \lambda_i \varepsilon_i + \dot{\varepsilon}_i, \ \lambda_i > 0, i = \{1,2\}, \quad (6)$$

where $\lambda_1, \lambda_2$ are positive constants to have stable sliding surfaces. Note that based on the defined sliding surfaces, the controller needs to regulate the combination of position errors and their time derivatives. The time derivative of the sliding surfaces is derived using Eqs. (1), (2), and (5):

$$\dot{s}_1 = \varphi_1 + \frac{1}{c_2}\varepsilon_2\omega^r - \frac{1}{c_1}v_x^r$$
$$\dot{s}_2 = \varphi_2 + \frac{x_0 - \varepsilon_1}{c_2}\omega^r. \quad (7)$$

In Eq. (7),

$$\varphi_1 = \left(-\frac{c_5}{c_2}v_x\omega - \frac{c_6}{c_2}\omega\right)\varepsilon_2 + \omega(-\omega\varepsilon_1 + v_x^d \sin\varepsilon_3 - \omega^d x_0 \cos\varepsilon_3 + \omega x_0 + \lambda_1 \varepsilon_2) - \lambda_1 v_x + \left(-\frac{c_3}{c_1}\omega^2 + \frac{c_4}{c_1}v_x\right) + \quad (8)$$

$$v_x^d[-(\omega^d - \omega)\sin\varepsilon_3 + \lambda_1\cos\varepsilon_3] + \dot{v}_x^d\cos\varepsilon_3 + \dot{\omega}^d x_0\sin\varepsilon_3 + \omega^d[x_0(\omega^d - \omega)\cos\varepsilon_3 + \lambda_1 x_0\sin\varepsilon_3], \text{ad}$$

$$\varphi_2 = -\omega(\omega\varepsilon_2 + v_x^d\cos\varepsilon_3 + \omega^d x_0\sin\varepsilon_3 - v_x) + (x_0 - \varepsilon_1)\left(-\frac{c_5}{c_2}v_x\omega - \frac{c_6}{c_2}\omega\right) + \dot{v}_x^d\sin\varepsilon_3 + (\omega^d - \omega)v_x^d\cos\varepsilon_3 - \dot{\omega}^d x_0\cos\varepsilon_3 + (\omega^d - \omega)\omega^d x_0\sin\varepsilon_3 + \lambda_2[(x_0 - \varepsilon_1)\omega + v_x^d\sin\varepsilon_3 - \omega^d x_0\cos\varepsilon_3].$$

The two control inputs $v_x^r$ and $\omega^r$ need to be designed to stabilize the defined sliding surfaces. The input controllers consist of two parts: The first controller part $\hat{*}^r$ is designed to maintain the errors on the sliding surface once the errors reach that surface (equivalent control term). The second control term $\bar{*}^r$ attempts to force the errors to reach the sliding surface starting from any combination of $\varepsilon_i$ and $\dot{\varepsilon}_i$ considering the uncertainties. Therefore,

$$v_x^r = \hat{v}_x^r + \bar{v}_x^r \\ \omega^r = \hat{\omega}^r + \bar{\omega}^r, \quad (9)$$

where

$$\dot{s}_2 = 0 \rightarrow \hat{\omega}^r = -\frac{c_2\varphi_2}{\hat{x}_0 - \varepsilon_1} \\ \dot{s}_1 = 0 \rightarrow \hat{v}_x^r = c_1\left(\varphi_1 + \frac{1}{c_2}\varepsilon_2\omega_r\right). \quad (10)$$

As was mentioned in Section II-B, $x_0$ is an unknown parameter in SSMR and is considered as a bounded parametric uncertainty $\hat{x}_0$ (i.e., $a \le \hat{x}_0 \le b$), which is based on the nonholonomic constraint of the robot [21]. The second control part is designed based on the general SMC convergence rule as follows ($k_i$ are positive constants):

$$\dot{s}_i \le -k_i sign(s_i). \quad (11)$$

As a result,

$$\bar{\omega}^r = -\left[\frac{c_2}{\hat{x}_0^{min} - |\varepsilon_1|}(-\bar{\varphi}_2 + k_2)\right]sign(s_2) \\ \bar{v}_x^r = -\left[c_1\left(\bar{\varphi}_1 + \frac{1}{c_2}|\varepsilon_2\bar{\omega}_r| - k_1\right)\right]sign(s_1), \quad (12)$$

where $\hat{x}_0^{min}$ is the minimum of $\hat{x}_0$ and $\bar{\varphi}_i$ is the maximum of $\varphi_i$ considering $\pm 25\%$ variation in the robot's dynamic parameters $c_{1:6}$ to consider system uncertainties. The inherent chattering problem is raised in Eq. (12) due to discontinuous sign functions. To alleviate this issue, the sign functions are replaced with *saturation* functions as follows:

$$\bar{\omega}^r = -\left[\frac{c_2}{\hat{x}_0^{min} - |\varepsilon_1|}(-\bar{\varphi}_2 + k_2)\right]sat\left(\frac{s_2}{\tau_2}\right) \\ \bar{v}_x^r = -\left[c_1\left(\bar{\varphi}_1 + \frac{1}{c_2}|\varepsilon_2\bar{\omega}_r| - k_1\right)\right]sat\left(\frac{s_1}{\tau_1}\right), \quad (13)$$

where

$$sat\left(\frac{s_i}{\tau_i}\right) = \begin{cases} \frac{s_i}{\tau_i}, & \left|\frac{s_i}{\tau_i}\right| < 1 \\ sign\left(\frac{s_i}{\tau_i}\right), & otherwise \end{cases}. \quad (14)$$

According to Eq. (14), the updated control terms in Eq. (13) behave similarly to Eq. (12) as long as $s_i$ is outside of the boundary layer (i.e., $s_i \ge \tau_i$). Therefore, the global stability of the closed-loop system needs to be investigated once $s_i$ gets inside the boundary layer. The result of the following stability analysis determines the thickness of boundary layers $\tau_i$. For that reason, first, we investigate if errors outside of the boundary layer get inside of it ($|s_i| \le \tau_i, |\varepsilon_i| \ge 2\tau_i$):

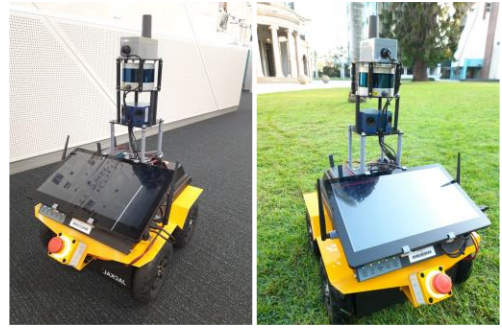

Figure 2. The Jackal platform in indoor (left) and outdoor (right) workspaces.

$$V_1 = \frac{1}{2}(\varepsilon_1^2 + \varepsilon_2^2) \rightarrow \dot{V}_1 = -(\lambda_1\varepsilon_1^2 + \lambda_2\varepsilon_2^2) + \varepsilon_1 s_1 + \varepsilon_2 s_2 \le -(\lambda_1\varepsilon_1^2 + \lambda_2\varepsilon_2^2) + |\varepsilon_1|\tau_1 + |\varepsilon_2|\tau_2 \le (1 - \lambda_1)\varepsilon_1^2 + (1 - \lambda_2)\varepsilon_2^2. \quad (15)$$

The above equation shows if $\lambda_1, \lambda_2 > 1$, then tracking errors get inside the boundary layer (i.e., $\dot{V}_1 \le 0$). The stability of the tracking errors now needs to be investigated once tracking errors get inside the boundary layer ($|s_2| \le \tau_2, |\varepsilon_2| \le 2\tau_2$):

$$V_2 = \frac{1}{2}(s_2^2 + \varepsilon_2^2) \rightarrow \dot{V}_2 = \frac{-\bar{\varphi}_2 + \bar{k}_2}{\tau_2}s_2^2 + \varepsilon_2 s_2 - \lambda_2\varepsilon_2^2 \le -\frac{\bar{K}_2}{\tau_2}s_2^2 + |\varepsilon_2||s_2| - \lambda_2\varepsilon_2^2 \le \\ -[|\varepsilon_2| \quad |s_2|]M\begin{bmatrix}|\varepsilon_2|\\|s_2|\end{bmatrix}, \quad (16)$$

$$M = \begin{bmatrix}\lambda_2 & -1/2 \\ -1/2 & \bar{K}_2/\tau_2\end{bmatrix}, \bar{K}_2 = -\bar{\varphi}_2 + k_2 > 0.$$

Choosing $det(M) \ge 0$ in Eq. (16) leads us to $\dot{V}_2 \le 0$. Therefore:

$$det(M) \ge 0 \rightarrow \tau_2 \le 4\lambda_2\bar{K}_2. \quad (17)$$

Finally, the same procedure for $\varepsilon_1$ is required to illustrate the stability of the tracking errors inside of the boundary layer ($|s_1| \le \tau_1, |\varepsilon_1| \le 2\tau_1$):

$$V_3 = \frac{1}{2}(s_1^2 + \varepsilon_1^2) \rightarrow \dot{V}_3 = s_1\left[\left|\frac{\varepsilon_2\bar{K}_2}{\hat{x}_0^{min} - |\varepsilon_1|}\right|\left(\frac{s_2}{\tau_2}\right)\left(1 + \frac{s_1}{\tau_1}\right) + \bar{K}_1\frac{s_1}{\tau_1}\right] + \varepsilon_1 s_1 - \lambda_1\varepsilon_1^2 \le \bar{K}_1\frac{s_1^2}{\tau_1} + \left|\frac{\varepsilon_2\bar{K}_2}{\hat{x}_0^{min} - |\varepsilon_1|}\right|\left(1 + \frac{s_1}{\tau_1}\right)s_1 + |\varepsilon_1||s_1| - \lambda_1\varepsilon_1^2 \le \left(\bar{K}_1 + \left|\frac{2\bar{K}_2\tau_2}{\hat{x}_0^{min} - |\varepsilon_1|}\right|\right)\left(\frac{s_1^2}{\tau_1}\right) + \left|\frac{2\bar{K}_2\tau_2}{\hat{x}_0^{min} - |\varepsilon_1|}\right|s_1 + |\varepsilon_1||s_1| - \lambda_1\varepsilon_1^2 \le \left(\bar{K}_1 + \left|\frac{16\bar{K}_2^2\lambda_2}{\hat{x}_0^{min} - |\varepsilon_1|}\right| + |\varepsilon_1|\right)\tau_1 - \lambda_1\varepsilon_1^2. \quad (18)$$

Choosing

$$\tau_1 \le \frac{\lambda_1\varepsilon_1^2}{\bar{K}_1 + \left|\frac{16\bar{K}_2^2\lambda_2}{\hat{x}_0^{min} - |\varepsilon_1|}\right| + |\varepsilon_1|}, \quad \bar{K}_1 = -\bar{\varphi}_1 + \bar{k}_1 > 0 \quad (19)$$

leads to $\dot{V}_3 \le 0$. The above stability analysis shows that choosing the boundary layer thicknesses in Eqs. (17) and (19) guarantee the global stability of the trajectory tracking system. Consequently, any $s_i$ inside the boundary layer asymptotically converges to the equilibrium point (i.e., $\lim(\lambda_i\varepsilon_i + \dot{\varepsilon}_i) \rightarrow 0$).

Additionally, [31] addresses the singularity problem in the controller arising from the denominator in Eq. (10), which is resolved by setting $\hat{x}_0 = 0$. However, their mathematical proof does not consider measurement noises. In this paper, we suggest using $\hat{x}_0^{min} - |\varepsilon_1|$ as the denominator in Eq. (10) to prevent singularity in the controller, accounting for measurement noises and camera corrections, which are common challenges in T&R systems.

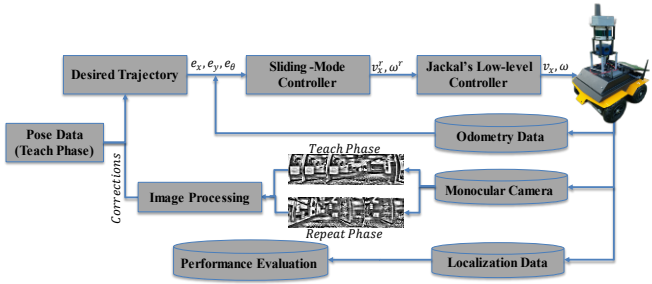

Figure 3. Proposed T&R navigation system pipeline: Teach and repeat phase images processed for odometry corrections and tracking error calculation. Then SMC generates robot velocity commands based on tracking errors.

Table 1. Hyperparameters of the image processing system [8] and the SMC.

| Image Processing parameter | Value |
|---|---|
| Image size | 115×44 |
| Patch normalisation size | 9×9 |
| NCC search range, $D$ | ±75 |
| Noise correlation threshold, $\bar{\rho}$ | 0.1 |
| Horizontal field of view, $deg$ | 75 |
| Distance threshold ($\tau_s$) | 0.1 |
| Angle threshold ($\tau_\theta$), $deg$ | 5 |
| $K_p, K_\theta, K_s$ | 0.01, 3 |
| **SMC Parameter** | **Value** |
| $x_0, \hat{x}_0^{min}, \hat{x}_0^{max}$ ($m$) | 0.05, -0.12, 0.15 |
| $\lambda_1, \lambda_2$ | 1.2, 2.6 |
| $k_1, k_2$ | 16.5, 20.5 |
| $\tau_1, \tau_2$ | 2.5, 3.5 |

Table 2. Indoor trajectory tracking. dist.: distance error. Errors: $m$ and $deg$ (± indicates standard deviation).

| Method | mae | | | Mean dist. | Stability |
|---|---|---|---|---|---|
| | X | Y | θ | | |
| Dall'Osto et al. [8] | 0.020 | 0.038 | 0.70 | 0.051 | Unstable |
| Proposed | 0.015±0.014 | 0.018±0.016 | 0.86 | 0.027 | Stable |

Table 3. Outdoor trajectory tracking. dist.: distance error. Errors: $m$ and $deg$.

| Method | mae | | | Mean dist. | Stability |
|---|---|---|---|---|---|
| | X | Y | θ | | |
| Dall'Osto et al. [8] | - | - | - | - | Unstable |
| Proposed | 0.028±0.023 | 0.032±0.028 | 0.74 | 0.046 | Stable |

## IV. EXPERIMENTAL SETUP

We assessed the T&R system using experiments indoors and outdoors with the Clearpath Jackal robot, conducted on carpeted floors at the QUT Centre for Robotics (Figure 2). Note that this laboratory space is shared with other researchers and activities, and there are moving people/objects in the environment. During the indoor testing, the robot was equipped with a Velodyne LiDAR to collect the robot's position as the ground-truth data using the HDL-Graph SLAM toolbox [41]. Jackal's default wheel odometry alongside a front-facing monocular camera were used to navigate the robot.

For outdoor testing, the robot was driven manually on uneven and wet grass at the QUT Garden Point campus (Figure 2), aided by an RTK-GNSS (Emlid Reach M2) with a 10 Hz sampling rate for outdoor localization. Identical manually tuned hyperparameters for image processing and SMC were applied in both indoor and outdoor tests (see Table 1). Repeat tests were conducted five times in both indoor and outdoor environments on the same day, albeit at different times. The overall proposed T&R navigation system pipeline is shown in Figure 3.

In both indoor and outdoor settings, the ground-truth data precision was approximately 1.5 $cm$ in stationary mode.

## V. RESULTS

This section presents the results of the proposed T&R navigation system in indoor and outdoor terrains, contrasting them with the state-of-the-art technique [8]. These experimental studies aim to illustrate the generality and the differences between systems with global stability versus previously used motion controllers, rather than solely comparing the accuracy of the trajectory tracking systems.

### A. Indoor Testing

In the indoor space, the robot was driven manually around the laboratory for the teaching phase, recording images and wheel odometry every 10 $cm$ or 5° of rotation. Each trial covered over 70 $m$ (20 tests, 1400 $m$ in total).

Figure 4 (left) shows the robot's trajectory in the teach and repeat phases in the global coordinate system. This figure shows that the proposed T&R system was able to successfully steer the robot and regulate the tracking errors regardless of the robot's linear velocity. Figure 4 (right) shows the distance error of the proposed and the previously developed technique by Dall'Osto et al. [8], with a comparison of the mean absolute error (*mae*) of $q^e$ and the mean distance error in Table 2. This table shows that the proposed method outperformed the previous method [8] with an average distance error of 2.7 $cm$ compared to 5.1 $cm$, while maintaining a stable closed-loop system and a 100% success rate in completing the manoeuvres.

To further demonstrate the stability of our method when adapting to varying manoeuvres, we altered the maximum linear commanded speed of the robot from 0.35 $m/sec$ to 0.6 $m/sec$ and repeated the experiment in the second trial. In this scenario, our proposed controller remained stable and closely followed the teach trajectory (Figure 4, left). However, the previous method [8] became unstable and uncontrollable, leading to premature termination due to safety concerns. The instability issue highlights that the previous method [8] was tuned for a specific task, lacking assurances for safe and successful task completion under changing conditions.

We next performed a sensitivity analysis to evaluate the success rate of the proposed controller when the hyperparameters $K_p, K_\theta, K_s, \tau_s, \tau_\theta$ of the path-planner are changed. To achieve this, three tests were carried out for each hyperparameter whilst changing one parameter at a time. During these tests, we changed the hyperparameters to $K_p, K_\theta = 0.005, K_s = 2, \tau_s = 0.2, \tau_\theta = 10°$ (see Table 1 for the default values) and the controller finished its task with 100% success rate. Notably, the adjustment of correction gains, i.e., $K_p, K_\theta$, directly impacts the motion controller's performance, as these parameters are designed to rectify the odometry drift issues.

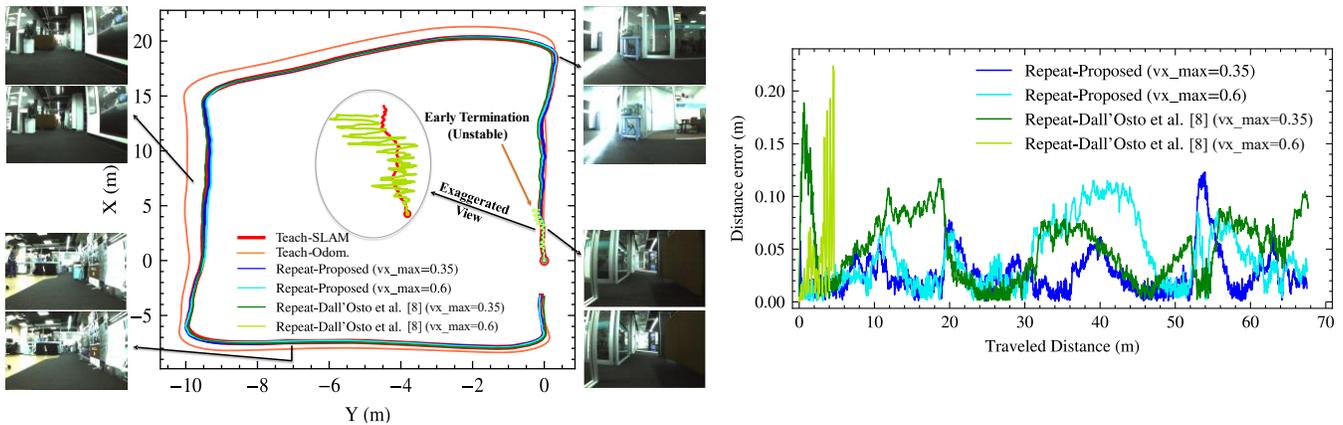

Figure 4. Indoor trajectory tracking results (left) and distance error (right). Example images of the teach (bottom image) and repeat (upper image) runs are shown to illustrate the variability of the workspace. Our proposed controller follows the teach trajectory with significantly smaller distance errors compared to [8], and successfully repeated the route even when nearly doubling the commanded speed of the robot, where [8] became unstable and needed to be terminated early. In the left figure, Teach-Odom denotes the robot's trajectory during the teach phase using odometry.

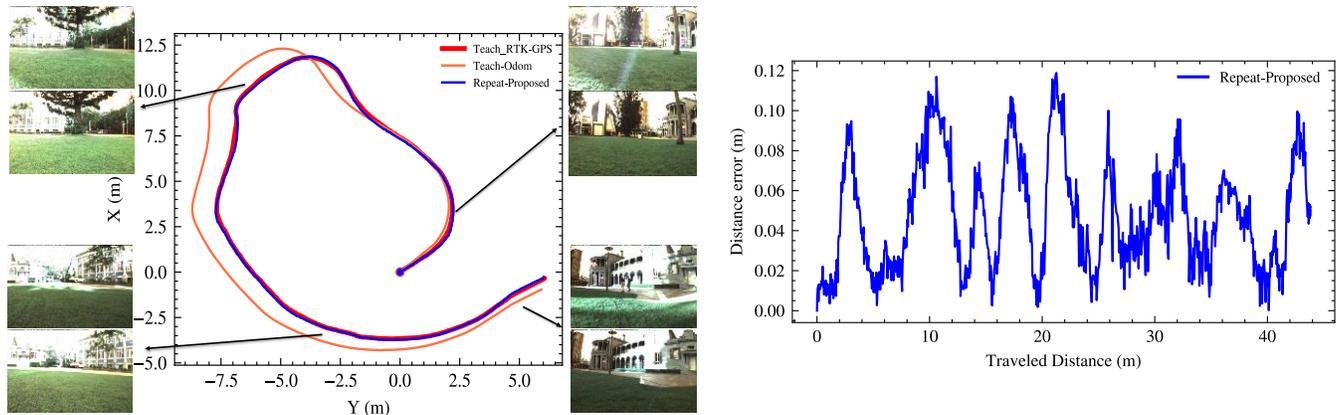

Figure 5. Outdoor trajectory tracking results (left) and distance error (right). Example images of the teach (bottom image) and repeat (upper image) runs are shown to illustrate the variability of the workspace and terrain conditions. The robot maintained a small error of less than 5 cm on average even under challenging terrain on uneven wet grass. In the left figure, Teach-Odom denotes the robot's trajectory during the teach phase using odometry.

## B. Outdoor Testing

In this section, the performance of the proposed technique is evaluated on an uneven, wet grass terrain. Similar to indoor testing, the robot was driven manually during the teach phase and the SMC was utilized with identical hyperparameters (Table 1) in the repeat phase. Note that we only evaluated the performance of the proposed method on off-road terrain due to safety reasons as the previous technique [8] already performed unstably in the less challenging indoor testing.

Figure 5 shows the robot's trajectory and the distance error over a 44 *m* path (5 tests, 220 *m* in total). Table 3 indicates that, on average, the robot maintained a 4.6 *cm* distance error while tracking the taught trajectory, achieving a 100% success rate in all five trials. Notably, this terrain causes a more challenging condition for the motion controller compared to indoor testing. Driving the robot on uneven wet grass can lead to an average of 25% slippage and 100 *mm/sec* undesired skidding [19], [20], which could induce greater odometry drifts as well as deviations from the desired path in both longitudinal and lateral directions. Figure 5 (left) shows that, during outdoor teaching, the robot's odometry displayed an average drift of 0.904 m, compared to 0.502 m indoors over the same distance, marking an 80.1% increase. This discrepancy underscores the increased challenges posed by outdoor terrains. Nevertheless, the robustness of our proposed method effectively managed this additional uncertainty caused by terrain conditions, ensuring system stability while guiding the robot.

## VI. CONCLUSION

In this paper, we proposed a novel robust T&R system based on a robust motion controller using the sliding-mode control technique. The proposed method guarantees the global stability of the closed-loop system considering unmodelled parameters and parametric uncertainties using low-cost and easy-to-integrate sensors, e.g., wheel odometry and a low-resolution monocular camera. The results show that the robustness of the proposed approach enables the T&R navigation system to operate in both indoor and off-road environments without the need for additional hyperparameter tuning to ensure stability, even on uneven terrains with varying traction levels.

In future work, we will investigate the impact of wheel-terrain interaction on our technique, as slip and undesired skid are primary contributors to wheel odometry failures and introduce uncertainty to the motion controller. Additionally, we plan to conduct large-scale outdoor testing across diverse terrains and enhance the robustness of our image processing system for improved performance under challenging lighting and sloppy terrains.